# Dynamic Systems Simulation and Control Using Consecutive Recurrent Neural Networks

Srikanth Chandar[1] and Harsha Sunder[2]

[1]Electronics and Communication Dept., PES University, Bangalore, India
Srikanth.chandar@gmail.com
[2]Microspin Machine Works Pvt. Ltd., Chennai, India
harsha.skillveri@gmail.com

**Abstract.** In this paper, we introduce a novel architecture to connecting adaptive learning and neural networks into an arbitrary machine's control system paradigm. Two consecutive Recurrent Neural Networks (RNNs) are used together to accurately model the dynamic characteristics of electromechanical systems that include controllers, actuators and motors. The age-old method of achieving control with the use of the – Proportional, Integral and Derivative constants is well understood as a simplified method that does not capture the complexities of the inherent nonlinearities of complex control systems. In the context of controlling and simulating electromechanical systems, we propose an alternative to PID Controllers, employing a sequence of two Recurrent Neural Networks. The first RNN emulates the behavior of the controller, and the second the actuator/motor. The second RNN, when used in isolation potentially serves as an advantageous alternative to extant testing methods of electro-mechanical systems.

## 1    Introduction

Electromechanical systems comprise actuators, controllers and motors: in practical field work, oftentimes it is not feasible to have access to these physical systems. We propose a novel approach to their simulation and control. Currently, the field of 'Industry and Automation' lacks a definite, robust and cost-effective model to perform testing of electro-mechanical systems, and its variants across plants.

Since the 1930's, control of these systems is traditionally done through the Proportional-Integral-Derivative (PID) controllers, which is in widespread use.[1,2]

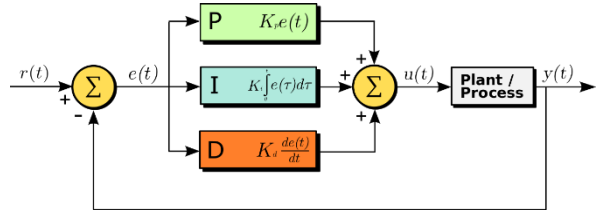

**Fig. 1.** PID Controller Equation

One of the most famous method of achieving control is using the Zeigler Nichols' methods. It is performed by setting the $I$ (integral) and $D$ (derivative) gains to zero. The "P" (proportional) gain, $K_p$ is then increased (from zero) until it reaches the ultimate gain $K_u$, at which the output of the control loop has stable and consistent oscillations.

$K_u$ and the oscillation period $T_u$ are used to set the P, I, and D gains depending on the type of controller used.

The correction $u(s)$ has the following transfer function relationship between error and controller output, in Laplace domain:

$$u(s) = K_p \left(1 + \frac{1}{T_i s} + T_d s\right) e(s) = K_p \left(\frac{T_d T_i s^2 + T_i s + 1}{T_i s}\right) e(s) \qquad (1)$$

The disadvantages of this technique are:

- It is time consuming as a trial and error procedure must be performed
- It forces the process into a condition of marginal stability that may lead to unstable operation or a hazardous situation due to set point changes or external disturbances.
- This method is not applicable for processes that are open loop unstable.
- Some simple processes do not have ultimate gain such as first order and second order processes without dead time.

There exist alternate methods like Tyreus-Luyben method, Damped Oscillation Method, etc. which aim to remove certain limitations of Zeigler Nichols' [3,4]; but not entirely. The reason being that all these methods have a framework, which on a fundamental level are simlar- to use a finite number of arbitrary constants which then encapsulates the entirity of a system and its behavior, regardless of the system's complexity level. The authors are of the opinion that (just) these constants are insufficient to completely describe any electro-mechanical system.

   İnstead of a controller being restricted to a fixed number of parameters, which then describes the system- the authors were inspired to look at the controller as a blackbox which builds a correlation between the inputs and the outputs of an electromechanical system, and thus computes the error and the suitable corrections. [5]

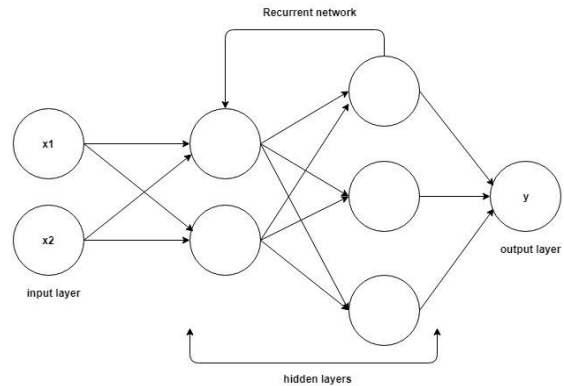

**Fig2.** RNN1 and RNN2 framework

Kwang Y.Lee [6] suggests that this blackbox can be achieved by the use of Diagonal Recurrent Neural Networks. The approach to use dynamic back propagation algorithms to these diagonal recurrent neural networks spurred the idea to use an enhanced version of the same.

It resulted in the creation of the Consecutive RNN approach, where the first RNN (RNN1) functions as an inverse of the second RNN (RNN2). RNN2 in isolation is the model which mimics the Electromechanical System (Microspin Machine [refer footnote]) to a high degree of accuracy.

## 2    Method

### 2.1    Machine, Model and Assumptions

The framework has two RNNs connected consecutively, and function as inverses of each other. The authors modelled an electro-mechanical system used in the textile industry by Microspin Machine Works[5] using RNNs.

The machine description is such that it takes in a time varying sequence of voltages, which is fed using Pulse Width Modulations (PWM(t)) directly proportional to the voltage at that instant. The output of this machine is corresponding time varying sequence of Revolutions Per Minute (RPM(t)).

The system in accordance to the laws of physics has an inertial lag during sudden spikes, or impulse PWMs. The prior knowledge that the authors had of the system before training the RNNs to replicate this model, is that this system has its own *flaws*- inertial lag, resonance, turbulence, lags at the start of steep function etc. The model built to replicate this machine, must have these *flaws* as well and must not be removed them in the name of efficiency.

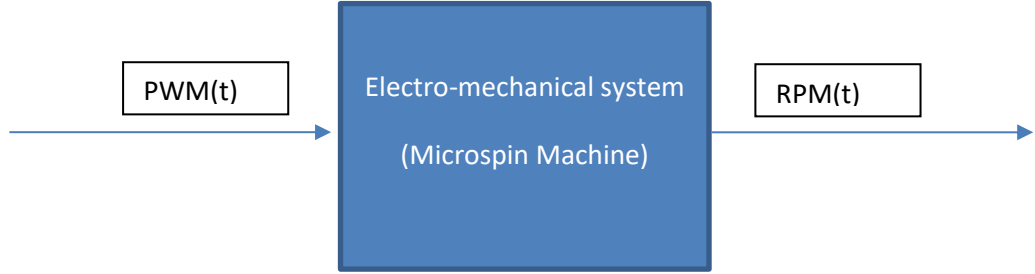

**Fig. 3.** Framework of Electro-mechanical System

     The machine has a device to capture the logged data, such as the PWM and RPM values at an instant. The logged data over a period of time had ~35,000 data points of the form (PWM,RPM). This data set comprised sinusoidal inputs, step inputs, impulse inputs and trapezoidal inputs (PWMs) - with varying slopes and peaks- and their corresponding outputs(RPMs).

     An assumption made at this point is that the system is symmetric about the origin; and hence doubled the data set we had from 35,000 to 70,000 by flipping signs. This assumption is valid, since logically a negative PWM would imply a negative RPM- meaning that the motor rotates in the opposite direction.

     This 70,000 data points is assumed to be a comprehensive list of details of the system, and used this as the training data to model the RNNs.

     After training the RNN (RNN2 henceforth) to mimic the machine to a high degree of accuracy, it is assumed that this model is a perfect simulation of the machine, and the controller RNN (RNN1 henceforth) is trained using the original dataset, and the data from RNN2.

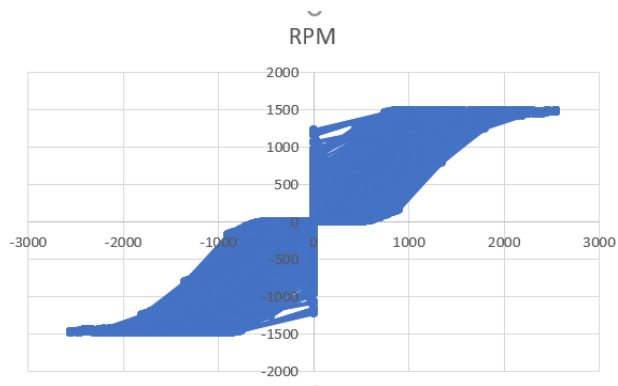

**Fig. 4.** Scatter plot of training data. (PWM vs RPM)

## 2.2    Problem Formulation and Analysis

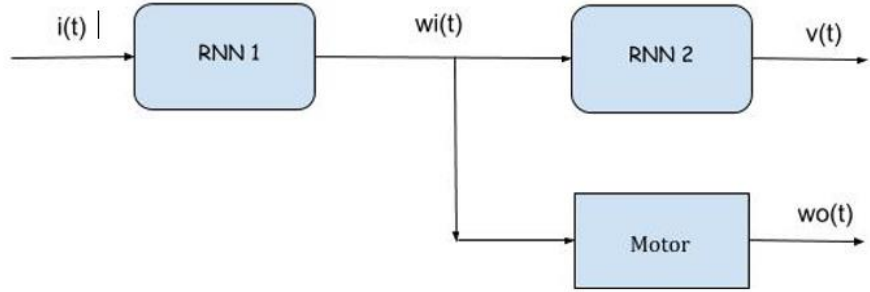

**Fig. 5.** Architecture of Consecutive RNN approach

Where:
$i(t)$ is Target RPM profile (and input to RNN1)
$w_i(t)$ is the Predicted PWM from RNN1 and the input to machine and RNN2
$w_0(t)$ is the Actual RPM of the machine
$v(t)$ is the predicted RPM of RNN2 (model of the machine)

$$\sum_t (\omega_o(t) - v(t))^2 \qquad (1.2)$$

$$\sum_t (v(t) - i(t))^2 \qquad (1.3)$$

**Fig. 6.** Equations to be minimized

RNN2 models the behavior of this machine to a high degree of accuracy by taking PWMs as inputs and output the corresponding RPMs. Post this, the physical machine is removed from the framework.

RNN1 models the controller by behaving as the inverse of the machine, i.e. to predict the voltage that has to be supplied to the machine in order to get a desired RPM profile.

The whole cycle is such: RNN1 takes in the Target RPM profile and predicts the PWM profile that needs to be sent into the machine (RNN2) to get back the same Target RPM Profile.

NOTE: This framework might resemble an Autoencoder, but the subtle difference is that, in this framework- the RNN2 mimics a real life system (the

electromechanical system). The output of RNN1 (or the input to RNN2) has a constraint that it should be such that it produces the Target RPM Profile, if given as inputs to the electromechanical system. This is not the case in the Autoencoder, as there is no constraint on the encoded pattern.

## 2.3    RNN2

- The machine logged data (PWM vs RPM), about 70,000 in number served as the training data (20% of which was used as test data)
- Given a time varying voltage, this model predicts the RPM at that instant of time
- Its built using LSTMs and combinations of dense layers to model this function of PWM vs RPM, as a one to one mapping.
  The training algorithm works to read 'x' time varying points of PWM and predict the 'x+1'th RPM. The models had varying step size (i.e. x=3, x=18 etc.). The most optimal model had x=3.
- The reason the model did not have x=1 (although its theoretically possible) is because 'x' cannot be too less; which would cause the model to not efficiently distinguish a data point when it has had two different histories.
- Even though RNNs account for the history through its memory feature, it was noticed that building batches of data and predicting the next in line output was more efficient. This is because the model now has a more unique data set (with a known history); so the predictions become more accurate.
- However, this batch size cannot be too large, as it would cause the system to lag – as it waits for as many intervals as the batch size, before predicting the next in line output.
- A fine balance between the two is the key

**Neural Architecture of a version of RNN2** .
Number of layers: 3 (2 LSTMs + 1 Dense)
Number of Neurons: 9
Number of Learnable Parameters: 122
Activation Function: Softmax
Optimizer: Adam
Loss function: Mean Square Error - between predicted output of RNN2 and true output of RNN2

$$MSE = \frac{1}{n} \sum_n (v_i - v_i')^2$$

Where:

$v_i$ is the ith predicted output
$v_i'$ is the ith true output
n is the total number of points

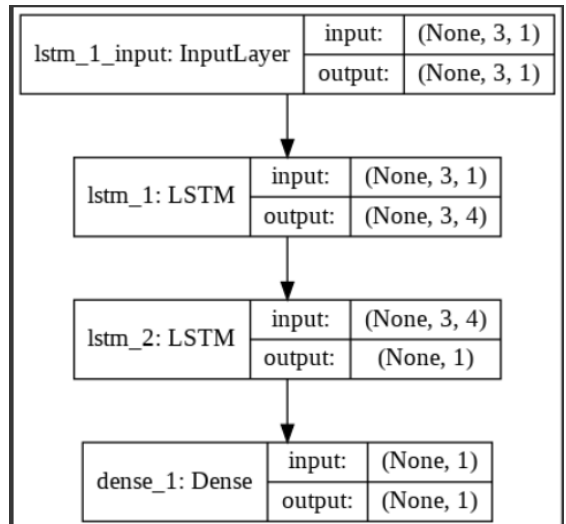

**Fig. 7.** RNN2 Neural Architecture

### 2.4 RNN1

- The input to RNN1 is the Target RPM Profile- at which we finally want the machine to run. We use the same training data used for training RNN2 but in an inverse fashion.
- Instead of predicting the RPM, given the PWMs; the model is trained to predict PWMs given a time varying RPM profile.
- This model is not simply an inverse of the RNN2. The method of training is completely different for the two models as they are fundamentally different in nature of data points.
- In RNN2, there was a one to one mapping of sorts between PWM and RPM, i.e. every PWM was unique and could have a corresponding RPM. The neural network was simpler, as it could uniquely relate a PWM to an RPM.
- However, with RNN1 the neural network has to be more complex, with more layers and had to be run on a greater number of epochs. This is because the RPMs were not all unique. Therefore, a prediction of an RPM, by simple logic is a one to many mappings.
- The modelling of RNN1 had to be done much differently for two reasons a)it is a one to many mapping, b) the error to minimized is :

$$MSE = \sum_t \quad (v_i(t) - i_i(t))^2 \qquad (2)$$

AND NOT:

$$MSE = \sum_t \quad (w_i(t) - i_i(t))^2 \qquad (3)$$

Where:

$v_i(t)$ is the ith predicted output of RNN2
$i_i$ is the ith input of RNN1
$w_i$ is the ith predicted output of RNN2
n is the total number of points

- Therefore, the loss function for RNN1 had to involve the output of RNN2 and the input of RNN1
- This is possible only if there were a custom loss function, or a way around that implies the minimization of the above error
- The custom loss function, took in the intermediate training outputs of RNN1 (PWMs) and fed it as inputs to already trained RNN2. The outputs of RNN2 – RPMs, were then used as parameters for the loss function along with the input to RNN1.
- The two parameters that the RNN1's loss function now depends on, are 1)input to RNN1 and, 2)The output of RNN2 when intermediate training outputs of RNN1 are fed as input.
- Note that the output of RNN1 is not directly a parameter for the loss function
- This custom loss can be also be achieved by connecting the RNN2, to the last layer of RNN1, and training RNN1 alone. (Since RNN2 weights are fixed to mimic the real electro-mechanical system)

**Neural Architecture of a version of RNN1.**
Number of layers: 4 (2 LSTMs + 2 Dense)
Number of Neurons: 52
Number of Learnable Parameters: 3,553
Activation Function: Softmax
Optimizer: Adam
Custom Loss function:  Mean Square Error - between output of RNN2 and input of RNN1

$$MSE = \sum_t \quad (v_i(t) - i_i(t))^2 \qquad (4)$$

Where:

$v_i(t)$ is the ith predicted output of RNN2

$i_i$(t) is the ith input of RNN1

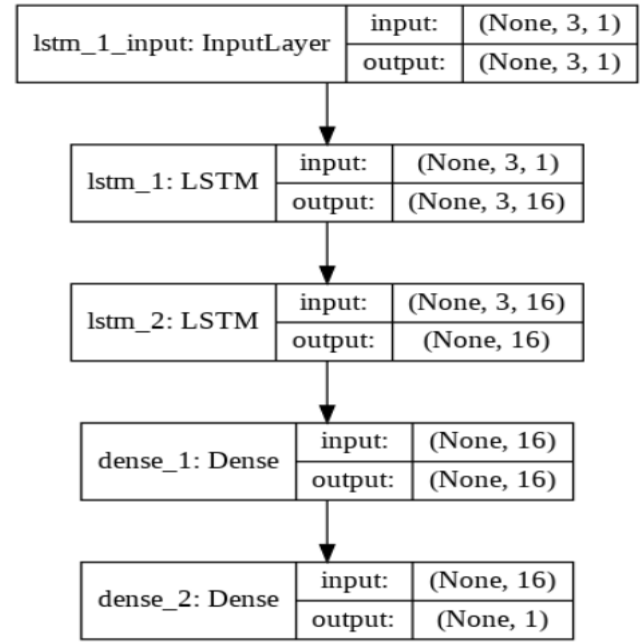

**Fig. 7.** RNN1 Neural Architecture

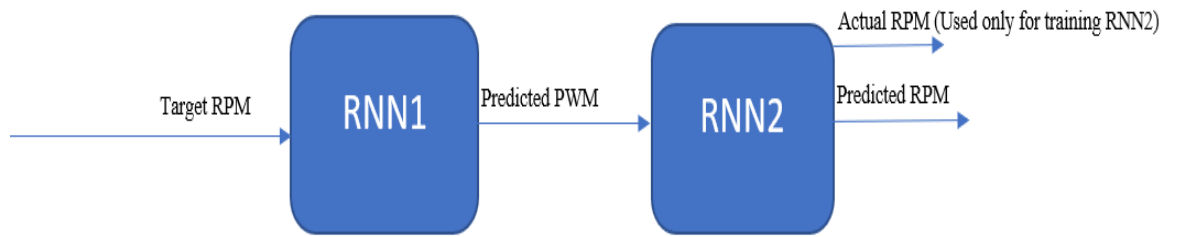

**Fig. 8.** Flow of Consecutive RNNs

## 3 Results

Using Consecutive RNNs in mimicking industry equipment, particularly in the electro-mechanical systems in the textile field, has been the primary focus of this paper, and has been achieved.

The mapping between Predicted and Actual RPM (in Fig. 9.) is almost spot on, with marginal error.

However, there is a significant time lag for the Predictions. The predictions are valid only after all the inputs of the step size have been fed to the RNN. There is an 18-unit time delay in the model with step size=18.

Fig.10. shows that this problem is solved by reducing the step size from 18 to 3.

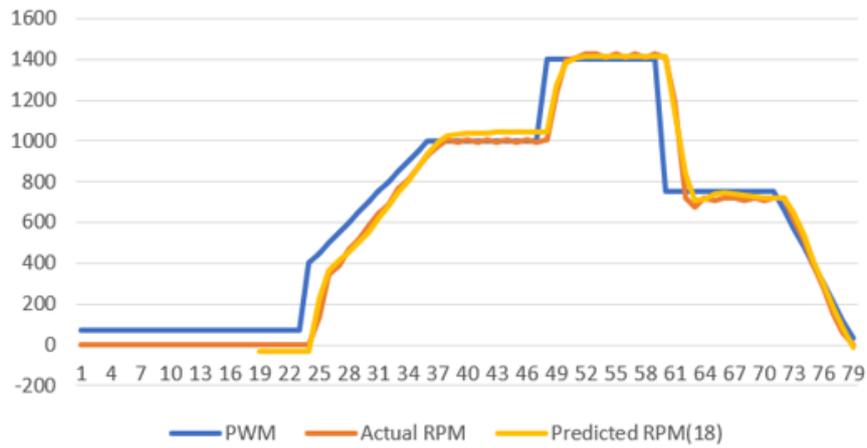

**Fig. 9.** RNN2 mimicking the motor almost identically; but with a lag. (x=18)

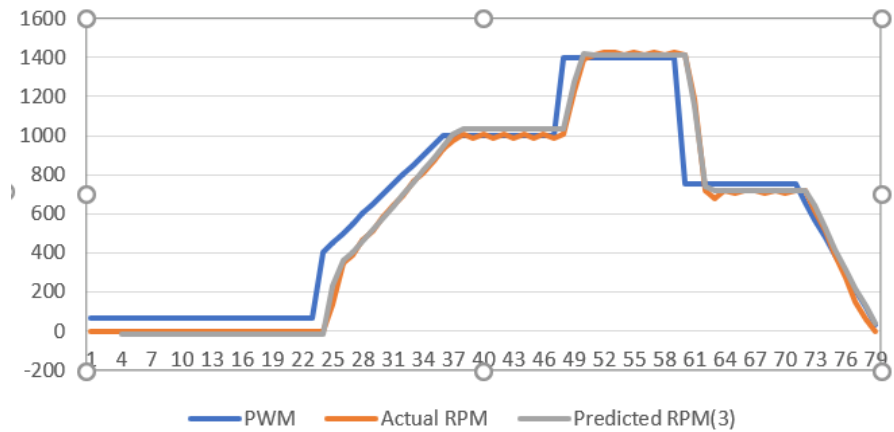

**Fig. 10.** RNN2 mimicking the motor almost identically; and with much lesser lag. (x=3)

The following graph (Fig. 11.) shows the relation between the predicted PWMs by RNN1 and the actual PWMs, when the input is a time varying RPM profile.
It shows how RNN1 behaves in isolation when step size=1
Here, the step size is 1. This has been done sine the 2 RNNs are connected, which would mean that any time lag would get doubled (one for each RNN).
It is noticed that this option of lag vs accuracy remains with the user, based on the situation. This lag can always be overcome by padding too.

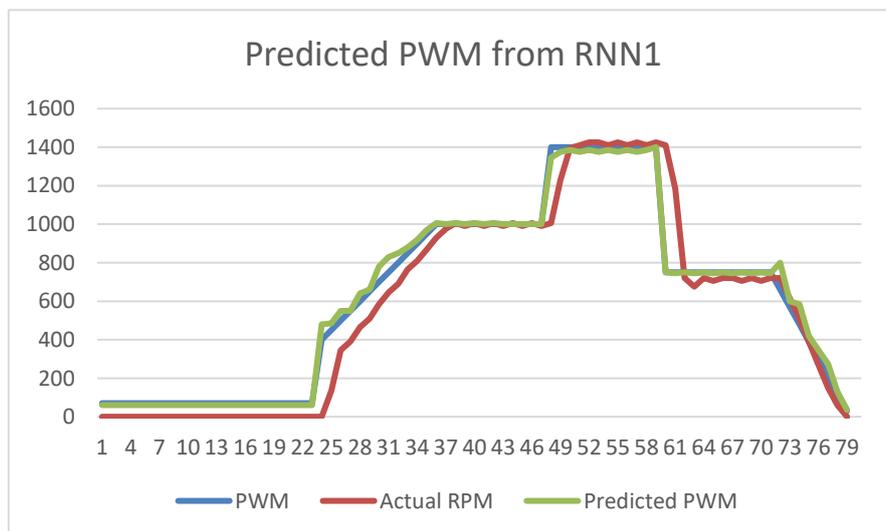

**Fig. 11.** RNN1 mimicking controller; and without lag. (x=1)

Fig. 12. and Fig. 13. capture the novelty of this paper, by comparing how the given electro-mechanical system behaves when controlled by the Consecutive RNNs, as opposed to classical PID control.

It is evident that the Consecutive Recurrent Neural Networks approach works as a better controller than the classical PID controller in case of this electro-mechanical system.

The difference margins between the output and the target is visibly lesser in the case of the proposed framework in comparison to the controllers used across textile industries.

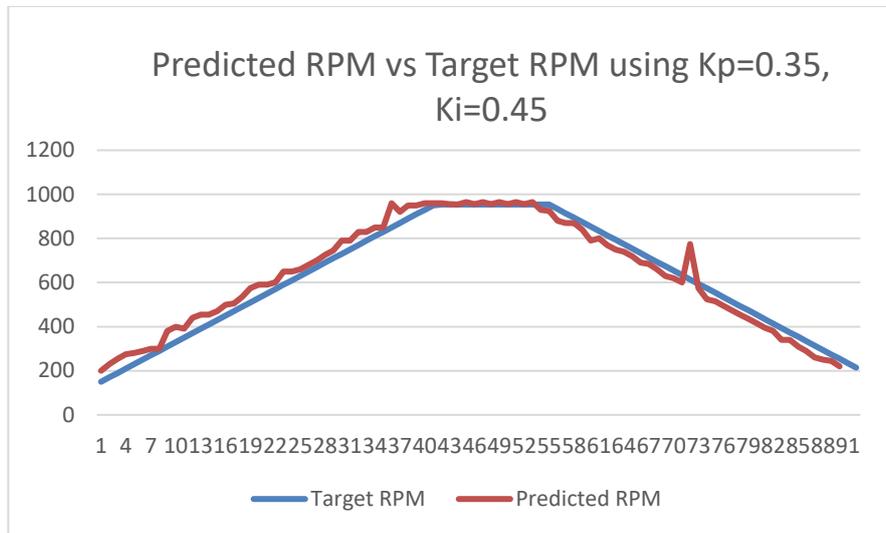

**Fig. 12.** PID contol using Kp, Ki constants which have been found using classical models involving trial and error.

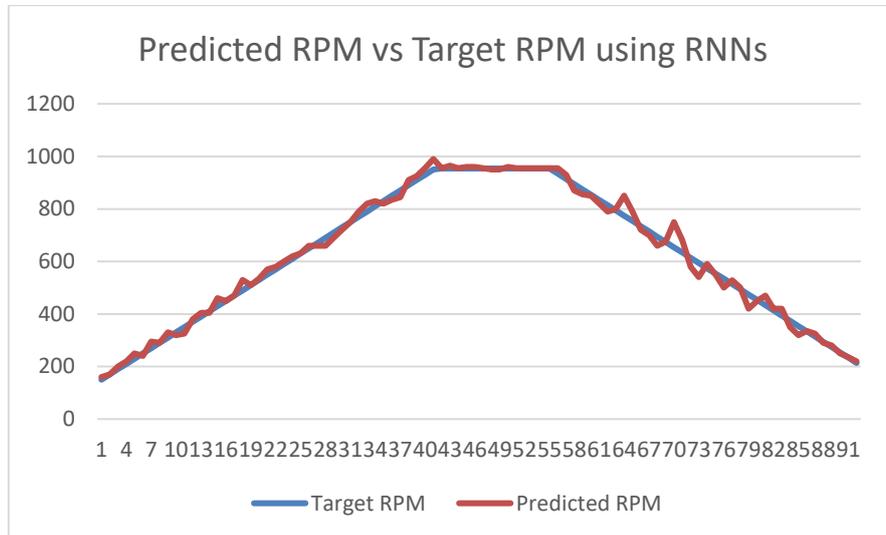

**Fig. 13.** Achiveing control using Consecutive RNNs

## 4    Conclusion

In conclusion, it is evident from the graphical representation of the results that the Consecutive Recurrent Neural Network approach a) works, and b) works better than the PID controller in certain cases such as - controlling the textile motor and machinery of Microspin Machine Works Pvt. Ltd.  The authors say certain cases and do not generalize this approach to be better than PID controllers as a whole just yet, as the analysis has been with only this class of machines. However, there a lot of future scope in this domain, as this might just be the start of a new branch of Control Theory.

The past efforts that do come close to the framework proposed in this paper, do not suggest with enough conviction that this type of controller actually works on a real time dynamic electro-mechanical system. [7,8]

The other conclusions to make from this analysis is that most electromechanical systems, even the nonlinear complex ones, can be modelled using RNNs, LSTMs etc. to a good degree of accuracy. A model has been built successfully, that could save Microspin Machine Works Pvt. Ltd. and the likes, from the cost of rent, power, etc. in the factories that are set up for testing (the electromechanical systems) purposes.

# 5    Acknowledgement

As Developers at Mircospin Machine Works Pvt. Ltd., we would first thank the company and all those associated in helping to make this project a success. From running around to log the data from the Machine, to technical coding needed in training the RNNs, it's been a collective effort of many in the workplace.

We thank Mr. L Kannan, CEO of Microspin Machine Works Pvt. Ltd., for being so closely related to a project with an intern, given his stature in the company. We also thank Dr. L Chandar (Support Vectors), and Mr. Asif Qamar (Support Vectors) who along with Mr. L Kannan, helped us in formulating the framework for this architecture. We also express our gratitude to Mr. Shritej Chavan (Student IIT-M), Mr. Pradeep Gopalakrishnan (Student IIT-M), Mr. Sumukh Nitundil (Student BITS Pilani), Ms. Shreya Vadrevu (Student PES University), Mr Chethan(Positive Integers Pvt. Ltd.), Mr. Sivaraman BV (Mircospin Machine Works Pvt. Ltd.), Prof. Rajini M(PES University) and Prof. Swetha R(PES University) for their support.

[i] Srikanth Chandar, 3rd year student B- Tech ECE, PES University, Outer Ring Rd, Banashankari 3rd Stage, Banashankari, Bengaluru, Karnataka 560085

[ii] Harsha Sunder, Product Architect, Microspin Machine Works Pvt. Ltd. 35, north street, Ganga Nagar, Sapthagiri Colony, West Jafferkhanpet, Chennai, Tamil Nadu 600083

Microspin Machine Works Pvt. Ltd. https://www.microspin.co.in